# Development of a hybrid machine-learning and optimization tool for performance-based solar shading design

Maryam Daneshi[1], Reza Taghavi Fard[2], Zahra Sadat Zomorodian[1], Mohammad Tahsildoost[1]

[1]*Shahid Beheshti University, Tehran, Iran*

[2]*University of Tehran, Tehran, Iran*

**Abstract**

Solar shading design should be done for the desired Indoor Environmental Quality (IEQ) in the early design stages. This field can be very challenging and time-consuming also requires experts, sophisticated software, and a large amount of money. The primary purpose of this research is to design a simple tool to study various models of solar shadings and make decisions easier and faster in the early stages. Database generation methods, artificial intelligence, and optimization have been used to achieve this goal. This tool includes two main parts of 1. predicting the performance of the user-selected model along with proposing effective parameters and 2. proposing optimal pre-prepared models to the user. In this regard, initially, a side-lit shoebox model with variable parameters was modeled parametrically, and five common solar shading models with their variables were applied to the space. For each solar shadings and the state without shading, metrics related to daylight and glare, view, and initial costs were simulated. The database generated in this research includes 87912 alternatives and six calculated metrics introduced to optimized machine learning models, including neural network, random Forrest, support vector regression, and k nearest neighbor. According to the results, the most accurate and fastest estimation model was Random Forrest, with an r2_score of 0.967 to 1. Then, sensitivity analysis was performed to identify the most influential parameters for each shading model and the state without it. This analysis distinguished the most effective parameters, including window orientation, WWR, room width, length, and shading depth. Finally, by optimizing the estimation function of machine learning models with the NSGA II algorithm, about 7300 optimal models were identified. The developed tool can evaluate various design alternatives in less than a few seconds for each.

Keywords: Solar Shading, Artificial intelligence, Optimization, Daylight, Glare, View, Machine Learning, Sensitivity Analysis

## 1. Introduction

The Indoor Environmental Quality (IEQ) impacts energy consumption and the performance of the buildings and users. This field has been widely studied by researchers in the last years. In general, environmental quality indicators include air quality, thermal comfort, visual comfort, and acoustic comfort [1]. The main activities in an office building depend on lighting quality, so visual comfort becomes very effective on users' performance. Visual comfort refers to areas such as daylight, glare, and view, so the building facade is the most critical factor related to these areas.

The envelope of a building is a significant element that affects energy consumption, to the extent that building facades are responsible for more than 40% of heat loss in winter and overheating in summer [2]. Components such as large glass windows or curtain wall systems, which are widely used in office buildings, can be significantly affected by direct solar radiation based on the orientation of the façade [3]. This makes them primarily dependent on solar radiation, which can lead to enormous cooling demands during hot periods. On the other hand, lighting energy has accounted for almost 19% of total electricity consumption [4], constituting over 20% of the consumed energy in office buildings [5], [6], which is a result of excessive radiation and glare. To avoid such glare, building users tend to close all blinds and rely on artificial lighting and mechanical air ventilation or air-conditioning, which contribute to a considerable increase in energy consumption [7]. Thus, solar shading devices become essential for optimizing and controlling solar radiation entering offices [8]. Besides that, in office buildings where users cannot freely adjust their position, optimal solar shading can create a productive work environment by improving the Indoor Environmental Quality. A study that explored the impact of external shading systems on cognitive task performance shows that higher performance is reported when the subjective sensation of visual discomfort is lower [9].

Recently, the majority of decisions in the design process of a building have been taken in the early design stage. Since the early design stage presents the greatest opportunity to obtain high-performance buildings, designers must be able to gather pertinent building performance information [10]. Shading systems play a fundamental role in the management of solar radiation, so selecting suitable screens should take place at an early stage of the design process since it significantly affects the energy balance of the building [11].

| Symbol | Description | Unit | Symbol | Description | Unit |
|---|---|---|---|---|---|
| ANN | Artificial Neural Network | - | Sh_depth | Shading depth | m |
| Area | Solar shading area | m2 | Sh_dis | Shading distance | m |
| ASE | Annual sunlight exposure | % | Sh_hdis | Shading horizontal distance | m |
| Avg_ill | Average illuminance | lux | Sh_mat | Shading material | - |
| Glass_vt | Glass visual transmittance | % | Sh_poh | Shading percentage of holes | % |
| HVC_60 | Horizontal visual cone_ 60 degree | % | Sh_rad | Shading radius | m |
| KNN | K Nearest Neighbour | - | Sh_tilt | Shading tilt | degree |
| MAE | Mean Absolute Error | - | Sh_vdis | Shading vertical distance | m |
| mDA | Mean Daylight Autonomy | % | SVR | Support Vector Regression | - |
| Obs_angle | Obstruction angle | degree | Win_height | Window height | m |
| PE | Probable Error | - | Win_num | Window number | - |
| $R^2$ | coefficient of determination | - | Win_side | Window orientation | - |
| RF | Random Forest | - | Win_sill | Window sill | m |
| RMSE | Root Mean Square Error | - | WWR | Window to wall ratio | % |
| sDA | Spatial Daylight Autonomy | % | X | Room width | m |
| Sh_angle | Shading angle | degree | Y | Room length | m |

Despite the significant impact that shading systems have on improving the performance of the building, they are mainly addressed in the detailed design stage or are not carefully modeled and evaluated in the early design stages. This issue questions the decisions related to some of the space and window parameters in the early stages. As a result, decisions in the field of shadings should be made along with other parameters related to space and windows. To achieve this, the user needs to evaluate and compare a wide range of design alternatives. Experts and companies cannot provide these facilities due to the time spent on simulation and the complex simulation process.

Therefore, in this study, a tool is presented that uses artificial intelligence and optimization to allow non-expert users to choose correctly from a wide range of alternatives and see the consequences of their decisions in the fastest time in the early stages of design.

The general method of this paper is to apply the variable parameters of space, window, neighborhood, and shading to an office room and simulate indicators in the field of daylight, glare, and view. In this way, first, the database was generated, in the next stage, it was developed with the help of machine learning algorithms, and the estimation functions were obtained. Also, the most effective parameters have been introduced by performing a sensitivity analysis step to add more capabilities to the tool. Finally, with the help of optimization algorithms, pre-prepared optimal models were collected to offer to the user.

This study focuses on external fixed shading systems. Static strategies are the ones that use environmental factors, such as wind and sunlight, to regulate the temperature of a building better. They require low maintenance, and they help to reduce energy consumption. Moreover, they do not need any help with additional mechanical systems, resulting in a convenient, economical way to achieve an energy-conscious building [12]. A simplified classification of the static solar shadings was presented by Lechner in 2014; shadings based on geometry mainly include overhangs, louvers, fins, eggcrate, and awnings; apart from the mentioned models, the vertical panels with different patterns are among the common shadings too.

The structure of this article contains Related Review Studies, Overview workflow, results obtained in each section, and the tool framework, discussion, and Conclusion.

## 2. Related review studies

Many studies have been done in the field of fixed shading design to evaluate a wide range of alternatives and select the best option.

A review of previous research in this paper has been done in two parts:
1. parametric design and optimization
2. use of artificial intelligence to evaluate the performance of shadings

In the first part of the study, the aim was to extract the variable parameters of the shading, performance evaluation metrics, simulation tools, and optimization algorithms along with its parameters. The second part of the review studies extracted machine learning algorithms and hyperparameters, inputs and outputs, validation methods, and indicators. There were research gaps in both sections that will be addressed.

*2.1. Parametric design and optimization*

The shading design method in this research is parametric design. Parametric design tools can easily help change variables and allow the user to visualize design variations almost in real-time, record and compare them.

Due to the time implications of revising all the model variations, optimization becomes a feasible option to generate all the iterative simulations in one single and more accurate process [12]. In a review study [13], among all 105 investigated studies, 34% of them analyzed the performance of various window and blind designs concerning visual comfort.

Manzan in [14], by applying overhangs for an office building with Variables such as installation height, depth, tilt angle, distance from the wall, and type of glass, evaluated the primary energy and $UDI_{100-2000\ lux}$ for two cities with different climates in Italy by optimization method. A different solution was found for each city to reduce the primary energy, which improved energy consumption by up to 19% for Trieste and 30% for Rome.

Gonzalez and Fiorito in [15], by optimizing parameters such as number, width, and tilt for the external louvers of an office building in Sydney, evaluated indicators such as DA and UDI and annual energy consumption. The results showed a save of 35% for annual energy consumption compared to the base scenario. The optimized solution consisted of 14 louvers with a slope of -10 degrees and a depth of 20 cm.

Manzan and Padovan in [16] obtained optimal models of overhangs for two exposures of an office building in Italy, with the following results: for the southern exposure, about 20% and for the southwestern exposure, about 23% decrease in energy consumption, and a UDI of about 90%. Also, with the use of external shadings, the operating hours of the interior blinds were considerably reduced by increasing the quality of daylight in the space.

Lee et al. [17] used the optimization and simulation methods to find the optimal shading design among models such as horizontal and vertical louvers, two models of vertical panels with geometric patterns, and random vertical and horizontal louvers. The variable parameters of this study include louvers tilt angle, number of louvers, louvers depth, number of the window divisions, and WWR. The output parameter was the percentage of space area with DF between 2 to 5%. Among the six types of shadings, vertical louvers reached the highest value (94%), and vertical panels got the lowest value (65%) by the simulation method. The values obtained by the optimization method were from 44 to 86%. The horizontal louvers had the highest value (86%), and the vertical panel had the lowest (44%).

Uribe et al. in [18] by designing Venetian blinds with slat angles of 0° and 45°, generic woven shade and perforated screen panels and considering variables such as glazing type, WWR, shading angle, etc. in three cities In Chile: Antofagasta, Santiago, and Punta Arenas, achieved results such as WWR values were obtained more than 90% because it increases daylight and this reduces lighting energy consumption. In Antofagasta, Single clear glazing was the best option, and in two other cities, double clear glazing was chosen. Optimal shading options were 45-degree louvers and generic woven shade because they provide visual comfort.

In [19], Ishac and Nadim identified a basic model, searched for an optimized solution, and validated the optimal solution, respectively. The following optimal alternatives were obtained by evaluating the light shelves and vertical fins through the optimization method: 1) In the south, an external light shelf of 120 cm, with a slope of 5 degrees upwards and an internal light shelf of 80 cm. 2) In the east, an external light shelf of 120 cm, an internal light shelf of 80 cm and vertical fins of 20 cm, and a -5 degrees angle with a distance of 40 cm from each other.

In studies, one or two shading models with limited variable parameters have been evaluated, and the parameters related to space have been assumed to be fixed, so it can be said that the results are not generalizable. As mentioned before, due to the importance of the shading parameters and their impact on other parameters of space, decisions regarding the parameters related to these should be made simultaneously and in the early stages of design. Also, in evaluating the performance of shadings, the view and cost analyzes have been left up.

*2.2. Artificial intelligence*

Estimation models based on machine learning algorithms are worth using by the building design community due to the lack of complex inputs, speed, low cost, and accurate predictions. Studies that have used these algorithms in recent years to predict the daylighting of spaces along with the parameters of the buildings under investigation, the algorithms used, the type of problems solved, the input parameters, outputs, and error metrics were reviewed by Ayoub in [6].

According to this research, the most used algorithms in previous studies are neural network algorithm (ANN), multiple linear regression (MLR), and support vector machine (SVM), and the most common problem types solved in the estimation models are regression, classification, and clustering, respectively. The most widely used output parameters are illuminance values, DA, sDA, and the most common error metrics are RMSE, PE, R2, and MSE, respectively.

These studies demonstrate that daylight and glare metrics can be estimated with acceptable accuracy by machine learning algorithms, but the approach to using these algorithms in articles was only to predict metrics and not to design a wide range of alternatives. Also, in most of these studies, variable parameters were mostly related to space, such as window dimensions, window orientation, space dimensions, obstruction dimensions, glazing parameters, etc. and in a few articles including [20], [21], [22], [23], [24] and [25] one or two typical shading models and a few numbers of variable parameters for each were assumed.

In a recent study by Lin et al. [26], machine learning models were used to design façades and predict daylight performance. For this aim, a vertical panel with various parameters such as Sky Exposure, Sky View, Visible Rate, Sunlight Hours, etc., was parametrically modeled. A database of 225 different façade conditions × 860 points on the test surface was generated. The daylight model was then trained using an artificial neural network and was capable of estimating the DA and ASE hours per grid with high accuracy with the test dataset (R2 value 0.91 and 0.88, respectively).

In another study by Nourkojouri et al. [27], a machine learning framework was developed to predict daylight and visual comfort metrics in the early design stages. A dataset consisting of 2880 alternatives was generated, and the parameters were related to the room, window, and shading. The outputs were useful daylight illuminance, spatial daylight autonomy, mean daylight autonomy, annual sunlit exposure, spatial-visual discomfort, and view quality. An artificial neural network algorithm was used, and the estimations' accuracy was predicted at 97% on average. A louver model was evaluated in this research, and the related parameters were limited.

## 3. Overview workflow

After reviewing the studies, the following pieces of information were obtained:

- variable parameters in categories of 1. space and neighborhood, 2. Opening, and 3. shading
- shading performance evaluation metrics in areas of 1. daylight and glare, 2. View, and 3. initial cost
- simulation software
- Machine learning algorithms as well as their parameters and validation methods

The main steps of this research are database generation, database development, sensitivity analysis, and optimization, respectively. Figure 1 provides an overview of the proposed method of this research and further information.

The main facility of the proposed tool is to predict the performance of the shading in the shortest time. Therefore, generating six databases for five shading models and the one without shading was necessary. After generating databases, machine learning models were trained; thus, estimation functions were obtained. Another facility of the tool is to introduce more effective parameters. As a result, a sensitivity analysis was performed. The most influential parameters on the performance evaluation metrics were identified, so the user would find a solution by changing only these parameters.

Finally, this tool would offer optimal shading models to the user. To achieve this facility, it was necessary to generate a database consisting of optimal models. This was possible by operating optimization on the estimation functions obtained from the machine learning procedure. For each case, a database composed of optimal models was collected; and unconventional models were eliminated.

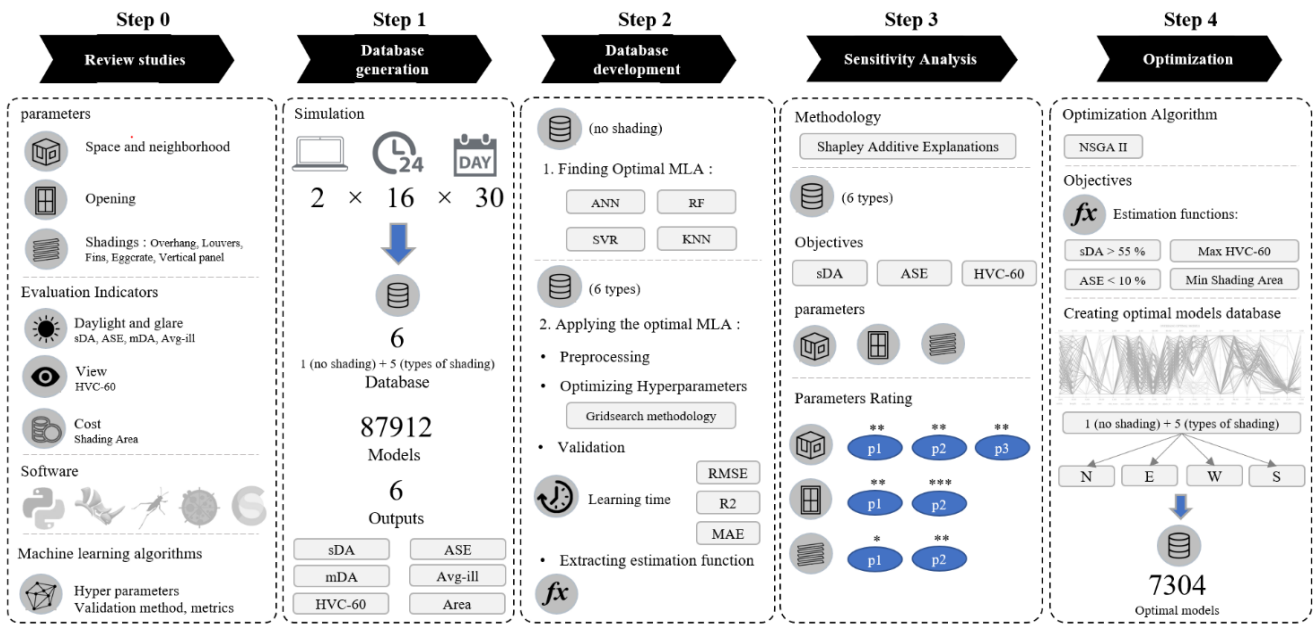

*Figure 1: Overview workflow*

## 3.1. Database generation

### 3.1.1. Input parameters

Several fixed and variable parameters have been assumed for the desired space, a side-lit office space in Tehran. The height of this space is equal to 3.5 meters, the length (Y) and width (X) of the space and the window orientation (win_side) are inconstant. Precisely opposite the opening, an obstacle whose length and width are equal to the main space is assumed. Still, the angle between the center of the opening and the upper edge of the obstacle (obs_angle) is unsteady. The average material reflectance values of interior surfaces are based on IES LM-83. The space is divided into two categories based on the window to wall ratio (WWR): ordinary and highly glazed spaces. The features of the window, including the window sill, the height of the window, and the number of windows, are different according to WWR.

Except for space-related parameters, five common models of shadings have been considered for the opening, which are overhangs, louvers, fins, eggcrate and vertical panels, each of which has some specific features that have been applied separately to the space. There are variable and fixed parameters that have been defined for each. A summary of variables and ranges is given in figure 2. The total number of alternatives is 87912, which table 1 shows these numbers for each shading model separately.

### 3.1.2. Outputs

Evaluation metrics in daylight and glare are sDA, ASE, mDA, and average illuminance. These have been selected due to the long-term evaluations, the representation of a single number for the whole space, and the acceptability in international certifications like LEED.

The view evaluation metric is the 60-degree horizontal cone (HVC-60). This metric is used to examine the interior to the exterior view, which shows the percentage of 360-degree horizontal band-limited at the top and bottom by 30-degree displacement of the horizon (derived from the human cone) for each point of a given plane and finally, the average number can be considered. Also, to assess the preliminary costs, the area of each shading has been calculated.

| Space and neighborhood | parameters | Range |
|---|---|---|
| | City | Tehran |
| | Building Type | Office |
| | X,Y (m) | (3,4), (6,7), (8,10) |
| | Height (m) | 3.5 |
| | Win_side | N, E, W, S |
| | Material Reflectance (%) | Floor: 20, Ceiling: 70, Walls: 50 |
| | Obs_angle (degree) | 0, 30, 60 |

| Opening | parameters | Range |
|---|---|---|
| Ordinary spaces | WWR (%) | 30, 60 |
| | Win_Sill (m) | 1 |
| | Win_height (m) | 2.2, 2.4 |
| | Win_num | 1, 2 |
| | Glass_vt (%) | 60, 80 |
| Highly glazed spaces | WWR (%) | 90 |
| | Win_Sill (m) | 0 |
| | Win_height (m) | 3.4 |
| | Win_num | 1 |
| | Glass_vt (%) | 60, 80 |

| Shadings | parameters | Range |
|---|---|---|
| overhang | Sh_dis (m) | 0, 1 |
| | Sh_depth (m) | 0.6, 0.8, 1 |
| | Sh_tilt (degree) | 0, 15, 30 |
| | Sh_Mat | Aluminium, wood |
| louvers | Sh_dis (m) | 0.4, 0.6 |
| | Sh_depth (m) | 0.2, 0.4, 0.6 |
| | Sh_tilt (degree) | 0, 15, 30 |
| | Sh_Mat | Aluminium, wood |
| fins | Sh_dis (m) | 0.4, 0.8 |
| | Sh_depth (m) | 0.2, 0.4, 0.6 |
| | Sh_angle (degree) | -15, 0, 15 |
| | Sh_Mat | Aluminium, wood |
| eggcrate | Sh_hdis (m) | 0.4, 0.8 |
| | Sh_depth (m) | 0.2, 0.4, 0.6 |
| | Sh_vdis (m) | 0.4, 0.8 |
| | Sh_Mat | Aluminium, wood |
| vertical panel | Sh_poh (%) | 0.3, 0.4, 0.5 |
| | Sh_dis (m) | 0.5, 1 |
| | Sh_rad (m) | 0.1, 0.2 |
| | Sh_Mat | Aluminium, wood |

*Figure 2: Input parameters ranges*

*Table 1: Number of alternatives*

| | No shading | overhang | louvers | fins | Eggcrate | Vertical panel |
|---|---|---|---|---|---|---|
| Number of alternatives | 648 | 23328 | 23328 | 23328 | 15552 | 1728 |
| **Total number of alternatives: 87912** | | | | | | |

*3.1.3. Software and tools*
Parametric modeling was performed in the Grasshopper Rhino plugin. In order to analyze daylight and glare due to a large number of simulation modes, the Climate Studio tool was selected, which is also on the Grasshopper platform and has acceptable speed and accuracy. Grasshopper Ladybug plugin was also chosen for visual analysis.

Regarding the input parameters of Climate Studio, except for the material reflectance values that have been mentioned before in figure 2, the height of the analysis surface is equal to the height of the users' desk in the office space (0.76 meters), and the grid size is 0.6 meters. According to the software's recommendation, the parameter ab was set in default mode, which is 6, and the number of samples for each sensor was considered 4096. Finally, the occupancy time of users was assumed to be 8 to 18 according to office hours.

Input parameters of the Ladybug plugin for view analysis include: The height of the analysis surface, which was assumed to be equal to the height of the user's eye in a sitting position is 1.2 meters, and the grid size is 0.6 meters.

After parametric modeling and applying the variable parameters, the simulations were performed parametrically with the help of the code written in the GhPython component. The total number of simulations is 87912 for each of the six metrics. The simulations and database generation took 30 days, and two systems were simultaneously processing 16 hours a day. After performing the simulations, the input parameters and output data were sorted, categorized, and prepared for computer programming.

*3.2. Database development*

*3.2.1. Machine learning*

The estimation models used in this study include Artificial Neural Network (ANN), Support Vector Regression (SVR), Random Forest (RF), and K-nearest neighbor (KNN). Each of these models contains hyper-parameters for which a specific range was initially considered, then by optimization using the Gridsearch method, which is based on examining the permutations of the parameters, Optimal hyperparameters were obtained.

In this study, the training data were normalized by the minmaxscaler method to train the estimating model better. This method helps the model to learn faster and more precisely. To validate machine learning models, the method of train test split has been used. In this way, the whole database was divided into 80% for training and 20% for testing, and then the training data was divided into 90% for training and 10% for validation. To prevent overfits, the learning model was validated once with 10% of training data (validation data) and once with 20% of total data (test data) that had not been used in any of the processing and optimization steps of the hyperparameters then the results were compared (figure 3). RMSE, R2, and MAE error metrics have been used to appraise the performance of the machine learning models.

Estimation models optimization and validation procedure should be done separately for each of the shading models and each of the outputs and four machine learning algorithms, so the total number of operations would be equal to $6 \times 6 \times 4, 144$ modes. Since optimizing hyperparameters with the Gridsearch method is very time-consuming due to time-saving, the four mentioned algorithms were examined only on the dataset without shading. The purpose of this is to find out the optimal algorithm, which means the most accurate (minimum error rate) and the fastest algorithm (minimum optimization time of hyperparameters) to be applied on other shading models.

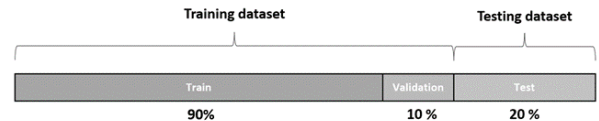

*Figure 3. Training, test and validation split*

*3.3. Sensitivity analysis*

There are various methods in evaluating the impact of parameters on outputs; the most common one is calculating the correlation coefficient with Pearson or Spearman methods. A method called Shapley additive explanations is one of the methods introduced in recent years that study the effect of different features on output data.

In this method, the algorithm predicts how important a model is by seeing how well the model performs with and without that feature for every combination of features. Then each feature's effect on the estimation process would be observed by Shap value. for additional information see [28].

*3.4. Optimization*

In this research, in order to accomplish optimization, a kind of genetic algorithm called 'elitist nondominated sorting genetic algorithm (NSGA-II)' has been used. This multi-objective algorithm with defective sorting is one of the most widely used algorithms in multi-objective optimization. Its main parameters include the population size and the number of generations which are assumed to be 500 and 20, respectively. For more information about the NSGA-II, please see [29].

## 4. Results

### 4.1. Simulation

According to the boxplots in figure 4, the average sDA for the state without shading is approximately 97% and for the states with shading varies from 68% (vertical panel) to 95% (overhang). There are some alternatives, especially in eggcrate and vertical panel datasets that sDA drops below 40%, so these cannot score in the LEED v4 certification.

The average ASE for the without shading dataset is 34% and for other datasets varies from 6% (vertical panel) to 26% (overhang). For louvers, eggcrate and vertical panel, it can be claimed that in half of the alternatives, this index is below 10% (recommended limit in LEED v4 standard).

In the absence of shadings and with the presence of overhangs, the average values of Avg_ill are more than 500 lux (office space requirement), so we can ensure that adequate light is provided, but the probability of glare increases.

The average values of HVC_60 are 7.4% and 3.1%-6.7% in alternatives without shading and with different shading models, respectively. This metric is calculated as a percentage of space, and the defined room is a side-lit space, as a result, its maximum value for the state without shading is 22.5%.

According to the results, the area varies from 0.79 to 104.5 meters, with the lowest values related to the overhang and the highest values related to the eggcrate.

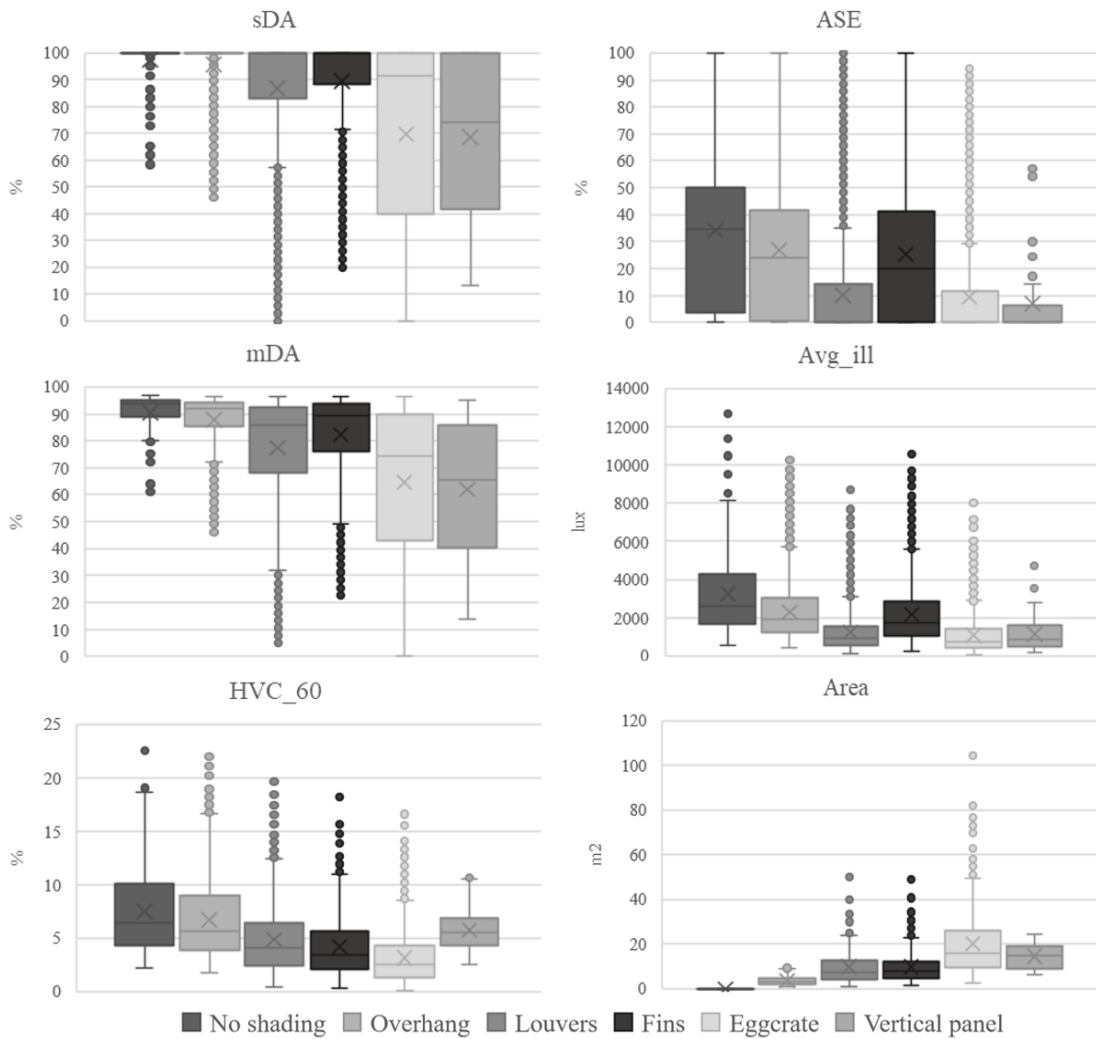

*Figure 4. Distribution of the Simulated outputs in the datasets*

## 4.2. Machine learning

### 4.2.1. Optimal estimation model selection

Optimization and validation for each of the datasets and each of the output metrics and four machine learning algorithms should be considered separately, so the total number of operations would be $6 \times 6 \times 4 = 144$. Since reaching optimal hyperparameters with the Gridsearch method is time-consuming, in order to save time, the mentioned algorithms were examined only on the database without shading. The purpose of this section is to find out the optimal algorithm which is the most accurate (minimum error rate) and the fastest algorithm (with minimum optimization time). Finally, the obtained model is set to be used for other datasets.

The results in table 2 indicate that optimization time according to the type of algorithm and output metric varies from 0.7 seconds to 7127 (About 2 hours). The minimum optimization times are related to the K nearest neighbor model, and the maximum times are associated with the artificial neural network. The optimal hyperparameters are also displayed in the table according to the type of estimation models.

The precision values for each of the outputs are shown in table 3. Concerning the random forest model, the mean absolute errors are below 0.5% for sDA, ASE, mDA, and HVC_60.

With the aim of selecting a model with the best performance, two metrics of average optimization time and the average MAE were established. For easier comparison, the obtained values were normalized by the minmaxscaler method. As shown in figure 5, the random forest model has had the lowest overall compared to other models.

### 4.2.2. Random forest model

This model was examined on all datasets, and an optimization and validation step were performed for each dataset, the results of which are expressed separately for each shading model in tables 4 and 5.

Regarding the overhang dataset, the maximum optimization time is for the average illuminance, which is 3664 seconds, and the lowest optimization time is for the sDA, 1806 seconds. The number of decision trees for different metrics varies from 1 to 465, and the hyperparameter max_depth for all metrics in the optimal state is 'none'. The RMSE for the average illuminance is 15.95, and for other indicators, it varies from 0 to 0.66. MAE is 0.24% for sDA, 0.06% for ASE, 0.12% for mDA, 9.4 lux for average illuminance, 0.02% for HVC_60 and 0 m2 for the shading area. In Figure 6, the difference between estimation and simulation results can be seen for different outputs. R2 for all outputs is in the range of 0.995 to 1. Learning time extends from 0.09 seconds for the area to 16.78 seconds for average illuminance.

Similarly, in tables 4 and 5, these values are reported for other datasets. In general, optimization time is almost below 1 hour, and learning time is below 17 seconds. Looking at the hyperparameters, the number of the estimation trees for some outputs like sDA and HVC_60 reaches over 400, despite other outputs like area and ASE, which in some cases 1 or 2 decision trees are enough to make an accurate estimation. Also, 'none' is optimal for the hyperparameter 'maximum depth' in all datasets. Results of optimal models in the validation section show that the r2_score differs from 0.995 to approximately 1. In almost all datasets, the MAE metric is below 1% for sDA, ASE, mDA, and HVC_60 and below 13 lux for the Avg_ill.

To ensure that overfit had not happened, the error rate was measured twice for the test and validation datasets, which by comparing these values, it can be observed that the difference is negligible.

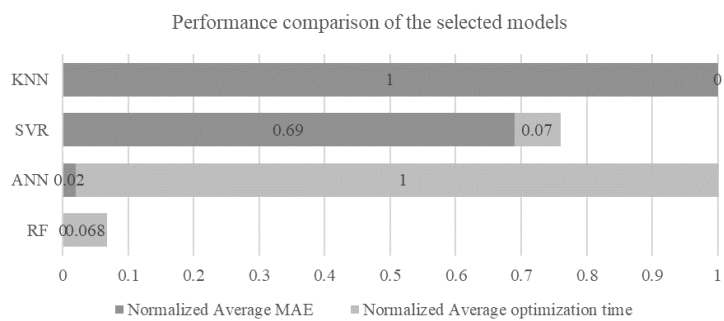

*Figure 5. accuracy and speed results of estimation models*

*Table 2: Optimization results of the machine learning models*

| Outputs | Hyperparameters | sDA | ASE | mDA | Avg_ill | HVC_60 |
|---|---|---|---|---|---|---|
| **ANN** | Optimization time (s) | 4078.120 | 2563.558 | 4184.448 | 7127.410 | 2278.613 |
| | Hidden layers | (330, 330, 330) | (70, 70) | (480, 480, 480) | (380, 380, 380) | (250, 250) |
| **SVR** | Optimization time (s) | 20.124 | 687.554 | 12.674 | 479.507 | 227.962 |
| | (C, gamma) | (500, auto) | (500, 0.5) | (67, auto) | (500, scale) | (186, 0.5) |
| **RF** | Optimization time (s) | 263.613 | 280.004 | 280.722 | 282.647 | 276.667 |
| | (Number of estimators, Maximum depth) | (45, none) | (179, none) | (264, none) | (38, none) | (90, none) |
| **KNN** | Optimization time (s) | 0.821 | 0.795 | 0.767 | 0.735 | 0.744 |
| | (Number of neighbours, weights) | (9, distance) | (4, distance) | (6, distance) | (4, distance) | (2, distance) |

*Table 2: Validation results of the machine learning models*

| | Validation metric | ANN | | SVR | | RF | | KNN | |
|---|---|---|---|---|---|---|---|---|---|
| | | Validation | Test | Validation | Test | Validation | Test | Validation | Test |
| **sDA** | R2_score | 0.964 | 0.984 | 0.232 | 0.290 | 0.988 | 0.993 | 0.473 | 0.462 |
| | MAE | 0.781 | 0.724 | 4.375 | 2.249 | 0.201 | 0.236 | 3.813 | 2.372 |
| | RMSE | 1.172 | 1.010 | 9.681 | 6.684 | 0.693 | 0.673 | 8.017 | 5.816 |
| **ASE** | R2_score | 0.999 | 0.999 | 0.968 | 0.981 | 1.000 | 1.000 | 0.695 | 0.683 |
| | MAE | 0.685 | 0.671 | 2.945 | 2.353 | 0.198 | 0.156 | 10.812 | 11.947 |
| | RMSE | 0.904 | 0.871 | 4.421 | 3.934 | 0.383 | 0.298 | 13.562 | 15.854 |
| **mDA** | R2_score | 0.986 | 0.993 | 0.728 | 0.771 | 0.999 | 0.999 | 0.633 | 0.678 |
| | MAE | 0.551 | 0.509 | 3.425 | 2.108 | 0.143 | 0.176 | 3.840 | 2.422 |
| | RMSE | 0.713 | 0.645 | 5.238 | 3.728 | 0.220 | 0.274 | 6.090 | 4.419 |
| **Avg_ill** | R2_score | 0.997 | 0.996 | 0.705 | 0.644 | 1.000 | 0.994 | 0.602 | 0.609 |
| | MAE | 97.894 | 79.711 | 624.676 | 713.795 | 31.839 | 54.72 | 859.484 | 994.942 |
| | RMSE | 141.350 | 134.175 | 939.372 | 1292.157 | 51.269 | 162.9 | 1090.799 | 1354.348 |
| **HVC_60** | R2_score | 0.995 | 0.984 | 0.940 | 0.925 | 0.992 | 0.967 | 0.826 | 0.869 |
| | MAE | 0.177 | 0.250 | 0.474 | 0.695 | 0.193 | 0.365 | 0.820 | 1.087 |
| | RMSE | 0.244 | 0.580 | 0.716 | 1.265 | 0.306 | 0.836 | 1.225 | 1.671 |

*Table 4: Optimization results of the optimal model (Random Forest)*

| Outputs | RF model | Overhang | Louvers | Fins | Eggcrate | Vertical panel |
|---|---|---|---|---|---|---|
| **sDA** | Optimization time (s) | 1806.143 | 2324.136 | 2437.453 | 1708.939 | 366.488 |
| | (number of estimators, maximum depth) | (413, none) | (420, none) | (465, none) | (116, none) | (39, none) |
| | Learning time (s) | 7.642 | 10.273 | 11.541 | 2.262 | 0.126 |
| **ASE** | Optimization time (s) | 2195.830 | 2010.405 | 2518.009 | 1277.760 | 269.236 |
| | (number of estimators, maximum depth) | (173, none) | (223, none) | (62, none) | (88, none) | (1, none) |
| | Learning time (s) | 4.517 | 4.453 | 1.678 | 1.063 | 0.003 |
| **mDA** | Optimization time (s) | 3244.606 | 3405.906 | 3294.459 | 2262.526 | 400.297 |
| | (number of estimators, maximum depth) | (352, none) | (192, none) | (190, none) | (167, none) | (136, none) |
| | Learning time (s) | 14.240 | 7.079 | 7.966 | 3.991 | 0.409 |
| **Avg_ill** | Optimization time (s) | 3664.601 | 3541.242 | 3505.211 | 2328.345 | 391.057 |
| | (number of estimators, maximum depth) | (236, none) | (334, none) | (344, none) | (316, none) | (310, none) |
| | Learning time (s) | 9.251 | 12.533 | 12.714 | 7.846 | 1.165 |
| **HVC_60** | Optimization time (s) | 3411.449 | 3185.859 | 3159.084 | 2041.512 | 373.341 |
| | (number of estimators, maximum depth) | (465, none) | (472, none) | (159, none) | (242, none) | (100, none) |
| | Learning time (s) | 16.784 | 15.135 | 5.400 | 4.871 | 0.298 |
| **Area** | Optimization time (s) | 2135.027 | 2141.359 | 2229.199 | 1644.668 | 317.747 |
| | (number of estimators, maximum depth) | (1, none) | (2, none) | (266, none) | (44, none) | (1, none) |
| | Learning time (s) | 0.094 | 0.089 | 6.116 | 0.832 | 0.007 |

*Table 3: Validation results of the optimal model (Random Forest)*

|  | Validation metric | Overhang | | Louvers | | Fins | | Eggcrate | | Vertical panel | |
|---|---|---|---|---|---|---|---|---|---|---|---|
|  |  | Validation | Test | Validation | Test | Validation | Test | Validation | Test | Validation | Test |
| **sDA** | R2_score | 0.995 | 0.996 | 0.996 | 0.996 | 0.995 | 0.995 | 0.996 | 0.997 | 0.997 | 0.997 |
|  | MAE | 0.281 | 0.244 | 0.687 | 0.687 | 0.644 | 0.621 | 1.200 | 1.123 | 1.019 | 0.905 |
|  | RMSE | 0.734 | 0.664 | 1.450 | 1.480 | 1.339 | 1.318 | 2.280 | 2.122 | 1.626 | 1.463 |
| **ASE** | R2_score | 1.000 | 1.000 | 1.000 | 1.000 | 1.000 | 0.998 | 1.000 | 1.000 | 1.000 | 1.000 |
|  | MAE | 0.062 | 0.067 | 0.054 | 0.051 | 0.099 | 0.115 | 0.060 | 0.056 | 0.000 | 0.000 |
|  | RMSE | 0.157 | 0.185 | 0.211 | 0.184 | 0.290 | 1.098 | 0.212 | 0.207 | 0.000 | 0.000 |
| **mDA** | R2_score | 1.000 | 1.000 | 0.999 | 0.999 | 0.999 | 0.999 | 0.999 | 0.999 | 1.000 | 1.000 |
|  | MAE | 0.129 | 0.127 | 0.321 | 0.318 | 0.293 | 0.296 | 0.561 | 0.586 | 0.326 | 0.300 |
|  | RMSE | 0.219 | 0.220 | 0.583 | 0.571 | 0.537 | 0.543 | 0.937 | 0.993 | 0.514 | 0.452 |
| **Avg_ill** | R2_score | 1.000 | 1.000 | 1.000 | 1.000 | 1.000 | 1.000 | 1.000 | 1.000 | 1.000 | 1.000 |
|  | MAE | 10.009 | 9.439 | 8.612 | 8.311 | 12.141 | 12.559 | 9.424 | 9.814 | 4.257 | 4.193 |
|  | RMSE | 17.024 | 15.95 | 15.632 | 14.29 | 19.391 | 20.261 | 15.615 | 16.485 | 6.579 | 6.305 |
| **HVC_60** | R2_score | 1.000 | 1.000 | 0.999 | 0.999 | 1.000 | 1.000 | 0.998 | 0.998 | 0.999 | 0.999 |
|  | MAE | 0.025 | 0.023 | 0.036 | 0.037 | 0.026 | 0.026 | 0.045 | 0.045 | 0.039 | 0.036 |
|  | RMSE | 0.046 | 0.042 | 0.091 | 0.089 | 0.043 | 0.043 | 0.103 | 0.094 | 0.055 | 0.057 |
| **Area** | R2_score | 1.000 | 1.000 | 1.000 | 1.000 | 1.000 | 1.000 | 1.000 | 1.000 | 1.000 | 1.000 |
|  | MAE | 0.000 | 0.000 | 0.000 | 0.000 | 0.000 | 0.000 | 0.038 | 0.030 | 0.000 | 0.000 |
|  | RMSE | 0.000 | 0.000 | 0.000 | 0.000 | 0.000 | 0.000 | 0.163 | 0.121 | 0.000 | 0.000 |

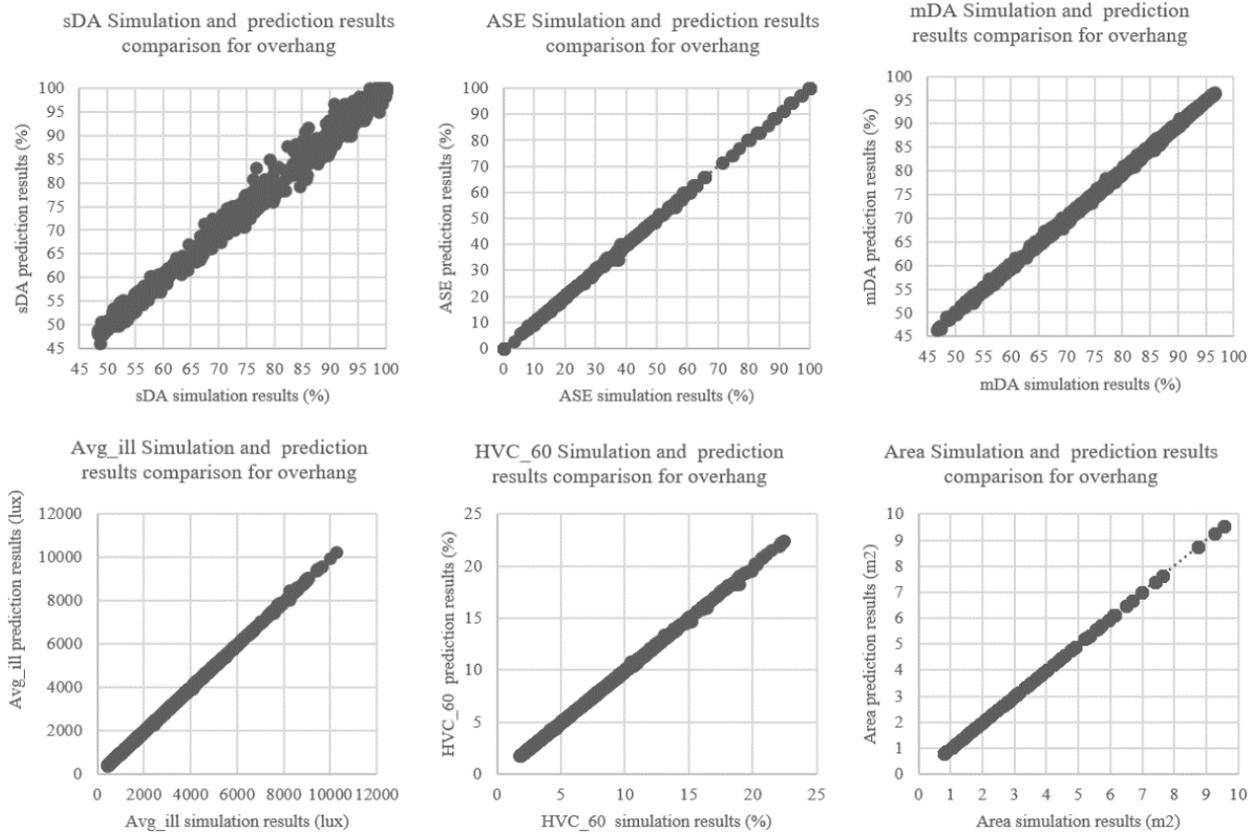

*Figure 6. Simulated vs. estimated outputs by Random Forest model for the overhang dataset*

*4.3. Sensitivity analysis*

In this section, several outputs, including sDA, ASE, and HVC_60, were selected to assess the effective parameters for each of the shading models. Due to the time-consuming calculation of the Shap value for each model, the sensitivity analysis was based on 200 randomly chosen alternatives from each dataset. It should also be noted that the assumed range of the input parameters in figure 2 directly impacted the results of this section.

In figure 7, the graphs indicate the average Shap value calculated for each of the parameters in all three outputs, and the sum of these three was considered as a metric for comparison.

As for most datasets, the most sensitive parameters are window orientation and WWR. Furthermore, for louvers and eggcrate, shading depth is one of the influential parameters on daylight and view metrics. For vertical panels, the distance of the panel from the façade has a remarkable effect on these metrics.

The following sensitive parameters are room length and width, shading distance, and shading tilt angle.

The shading material is in the last place in almost all datasets. As a result of this study, there is not much difference between wood and aluminum.

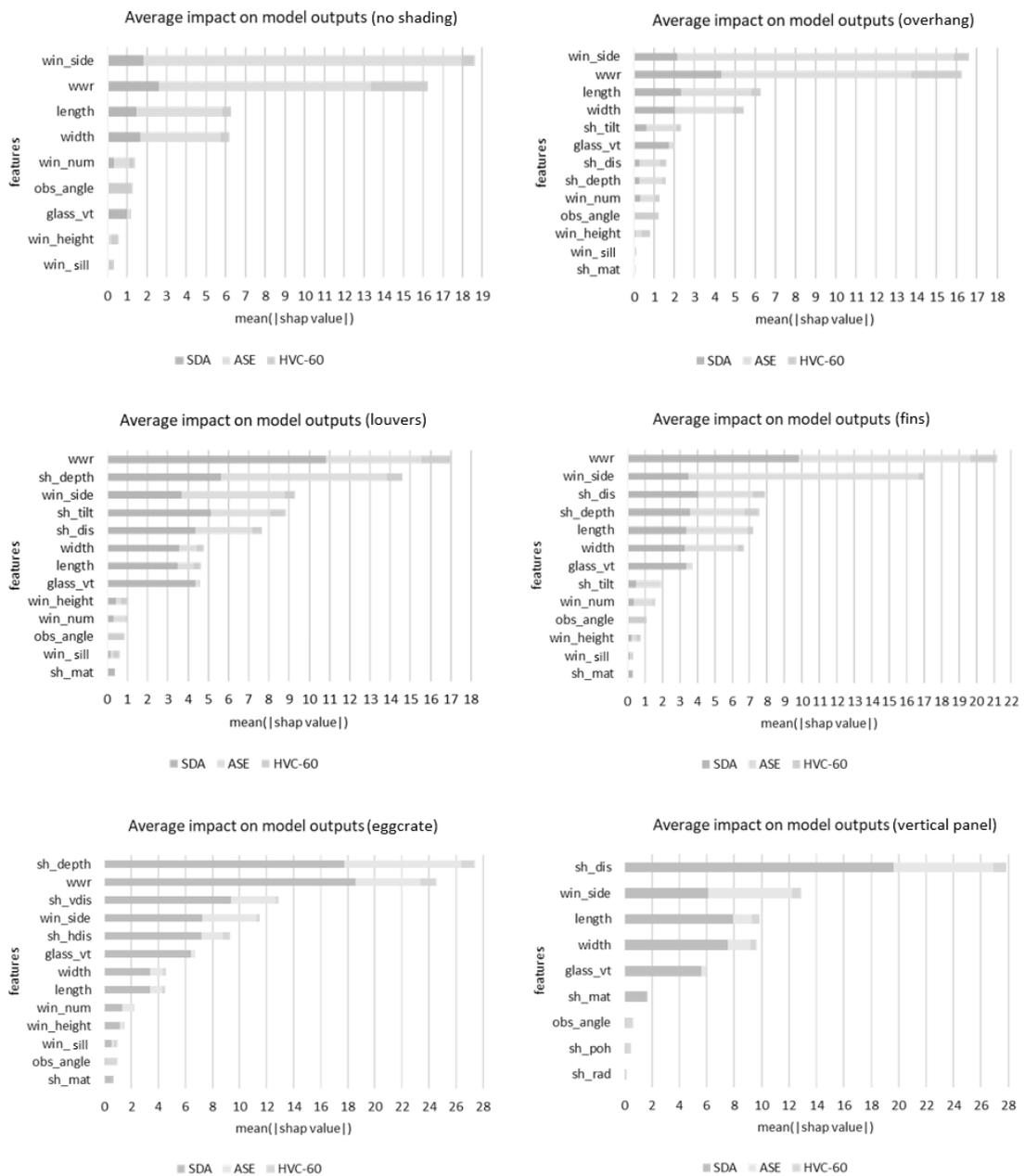

*Figure 7. Overall sensitivity analysis of the output metrics*

The rankings of these parameters in each database are different, which can be seen in more detail in figure 7.

In general, the impact of shading parameters in those shadings that cover a larger area of the opening, such as louvers, vertical panels, and eggcrae, tend to be even more significant than the parameters related to the space and opening.

*4.4. optimization*

Optimization was performed to achieve the best design parameters for four different window orientations separately using the NSGA II algorithm. The estimation functions that had been acquired from the machine learning section of 4 selected metrics, sDA, ASE, HVC_60, and shading area, were optimized in order to find solutions with maximum daylight and view and minimum glare and cost. As a result, for each shading model, a total of 2000 optimal models were obtained. Some of the detected models were unconventional; for example, with narrow windows, a Python code was written to solve this problem, and these models were removed.

Finally, a database of 7304 optimal models was generated, the specifications of which are shown for each of the shading models in figure 8. Each color in the graphs represents a window orientation.

As can be seen, the optimal models cover a wide range of alternatives. As a result, the users would be able to find the optimal shading for the various rooms by entering some effective parameters such as the length and width of the space and the window orientation. For further explanation, the more details the users enter, the fewer alternatives, but much more closer to their case would be offered.

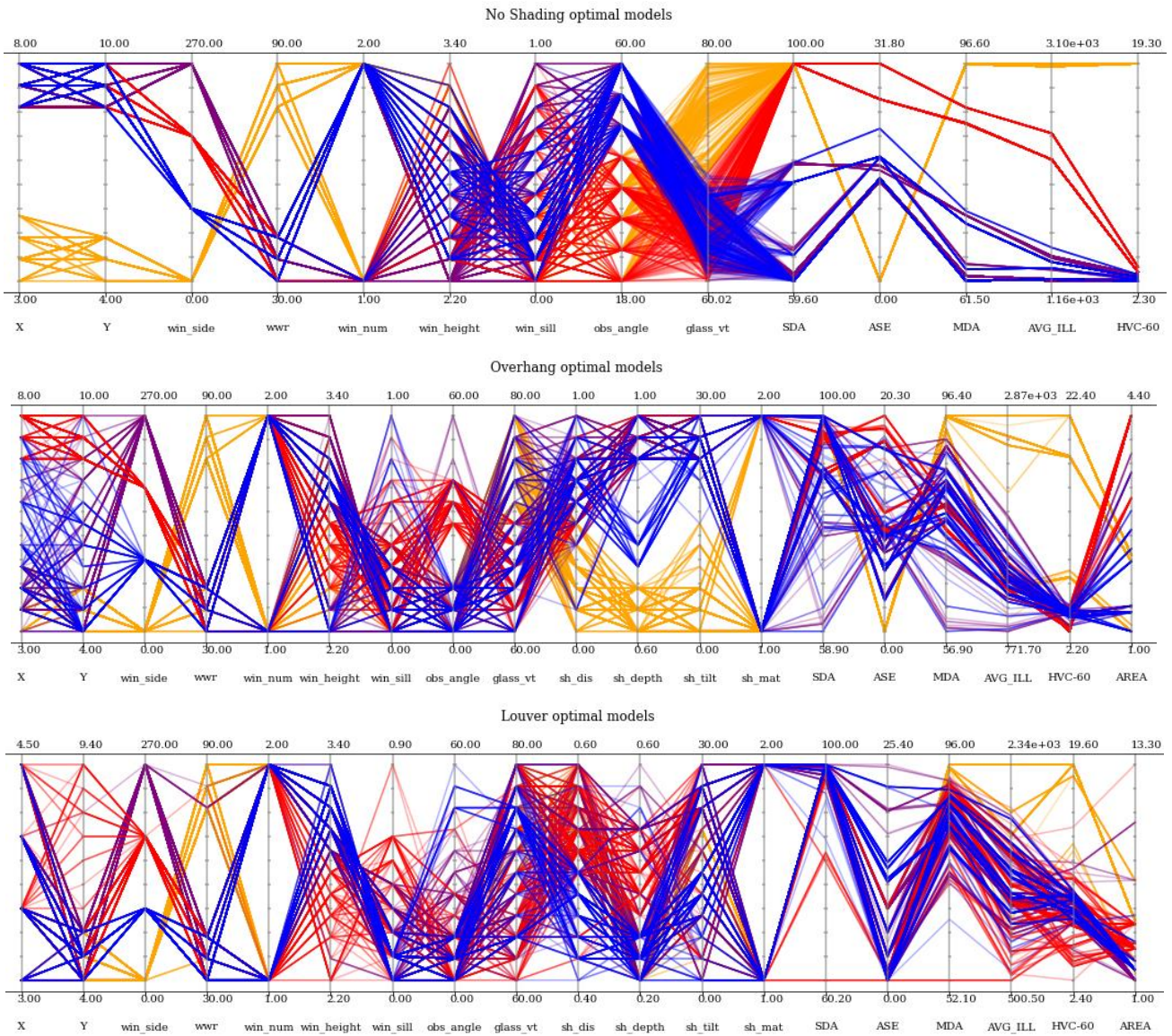

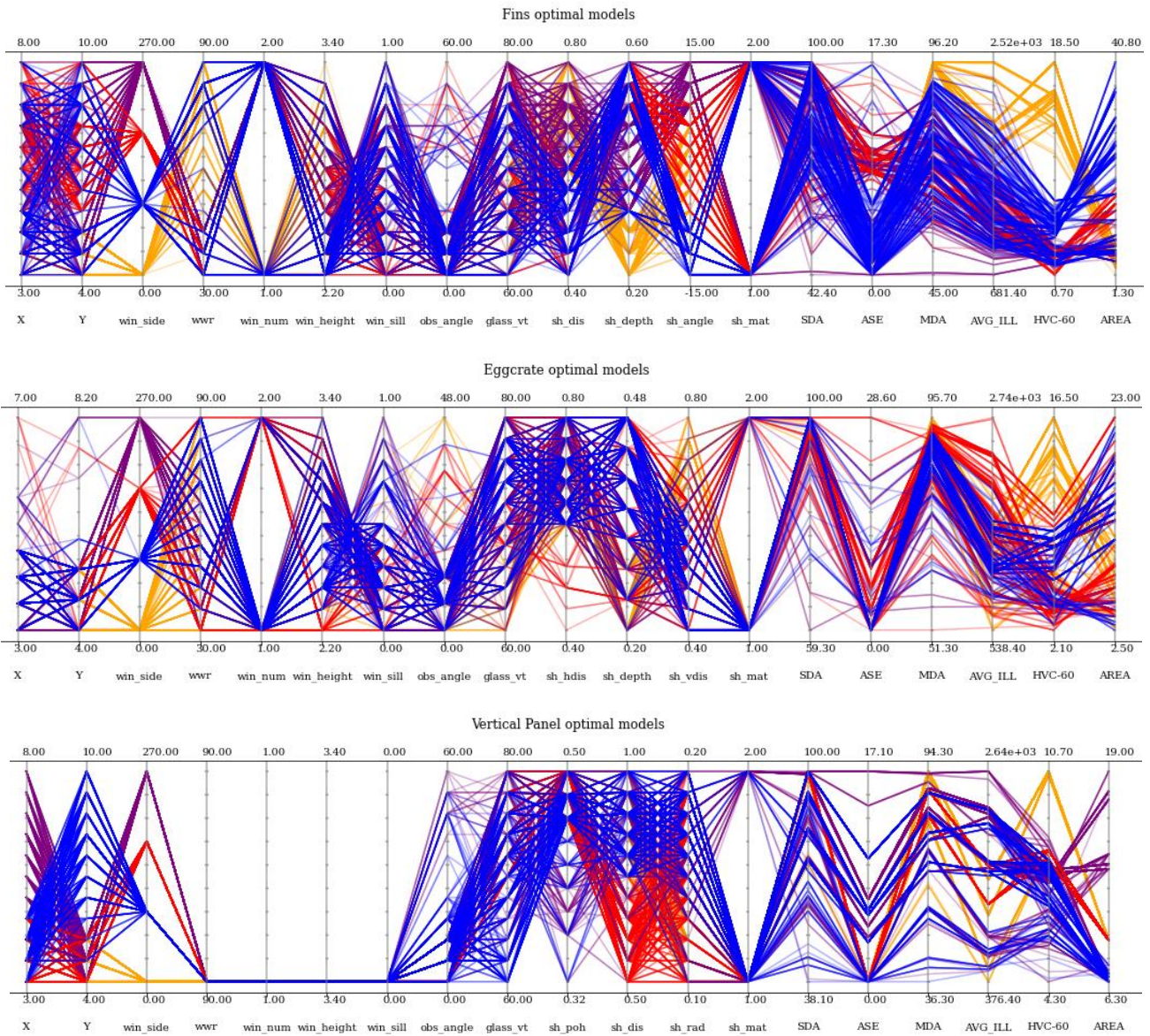

*Figure 8. Specifications of 7304 optimal models identified by the NSGA-II algorithm*

*4.5. Tool framework*

The proposed framework would consist of two main parts. Initially, by selecting the shading model and even the state without shading, the user would be able to see the parameters labeled with '*' based on their impact on the outputs. This would help the users to enter sensitive parameters more accurately.

By entering all the parameters, the user in the first part can view the estimation results by the machine learning model, and by entering a limited number of parameters, the tool has the ability to provide some optimal models with their results to offer the user.

Also, in addition to the outputs, the LEED v4 score was calculated for daylight so that the user can evaluate the alternative with international certificates.

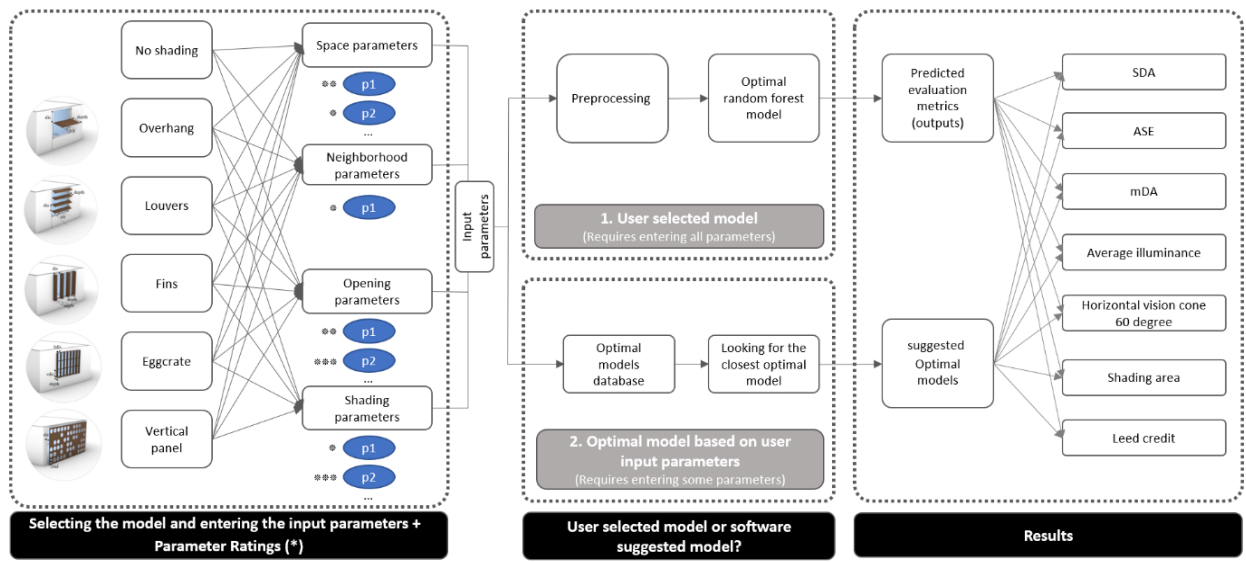

*Figure 9. Proposed tool framework*

## 5. Discussion

This research shows that evaluation and decision-making among a vast number of design alternatives that were previously challenging and costly can be quickly done by artificial intelligence and optimization methods. Shading design can be a challenging task due to different conditions of space, opening, and neighborhood. The proposed tool would give the user the ability to make decisions related to these in the early stages of design with acceptable accuracy and high speed. To achieve this, a total of 87,912 different alternatives were introduced to machine learning algorithms, and the learning model was able to estimate the outputs with high precision. Due to the diversity of parameters, the parameters with a more significant impact on the outputs were obtained by performing a step of sensitivity analysis. In the next step, a database of 7304 optimal models was generated by optimizing the outputs estimation functions. Finally, the developed tool would be able to evaluate the performance of the user-selected model or suggest optimal pre-prepared models. However, there are limitations in this study together with recommendations for future research, which will be addressed in the following.

- In this study, single office space in Tehran with several variable parameters based on review studies was assumed; however, many parameters were not addressed in this study. For example, the space can be studied in different climates or with two or more window orientations. Also, more diverse models of external shadings, or the internal ones like blinds, can be considered. In addition, the materials used in this research were the standard materials suggested by the guidelines, while the material reflectance value has the potential to be an effective parameter. As for the neighborhood, only one building in front of the opening was modeled; this scale can be larger.
- Regarding outputs, ASE was used to evaluate glare in this study. Other visual comfort metrics, especially for more accurate glare assessment such as DGP, can also be calculated. The DGP metric can be assessed for hours of the year with the maximum illuminance level on each orientation. It would also be useful to provide images of the interior. Additionally, Shading design has a significant impact on energy consumption, especially cooling and lighting energy, so calculating energy-related metrics can help the users make better decisions. Other performance comparison metrics can also be used for easier decision-making, for example, the percentage of improvement for sDA or ASE compared to the state without shading.
- As for generating and developing the datasets, simulation and optimization methods have been used in this study. For this purpose, CVAE and GAN networks, which are constructive neural network models, can be used to expand the database.
- There are different ways to validate a machine learning model, and this research used the method of train and test split. In addition, a new database can be defined outside the presumed range for each parameter, and the accuracy of the learning model in estimating outputs can be measured. The assumed range for the parameters can be extended if acceptable accuracy is achieved.

## 6. Conclusion

A hybrid machine learning and optimization tool framework has been developed in this research which can be used in the design of static solar shadings and evaluation of their performance. Design parameters were related to the shading, space, and opening because these are integrated, and evaluation metrics were in the field of daylight, glare, views, and initial costs. These outputs have been simulated and estimated for five different shading models applied to a single shoebox space. The machine learning results indicated that all the metrics can be predicted by the optimal estimation model, which was the random forest model, with high accuracy and pace. Aiming to enhance the tool further, the optimization of the estimation functions, which had been obtained in the machine learning section, led to a large number of optimal models, and by generating a database consisting of these optimal models, the user would be able to find the solution with no challenge at all. This procedure would form a tool capable of overcoming the present costly and time-consuming methods and can be used for designing and evaluating a majority of solar shading alternatives for different spaces in the early stages of design.